\documentclass[journal]{IEEEtran}
\ifCLASSINFOpdf
  \usepackage[pdftex]{graphicx}
\else
\fi
\usepackage{textcomp}
\usepackage[gen]{eurosym}
\usepackage{array}
\usepackage{amsmath}
\usepackage{amsfonts}
\usepackage{amsmath,amsfonts,amssymb}
\usepackage{multicol}
\usepackage{multirow}
\usepackage{bbding}

\begin{document}
%
\title{Spatial Dual-Modality Graph Reasoning for Key Information Extraction}
%
%
%

\author{Hongbin Sun,
        Zhanghui Kuang,
        Xiaoyu Yue,
        Chenhao Lin and Wayne Zhang
\thanks{Hongbin Sun, Zhanghui Kuang, Xiaoyu Yue, and Wayne Zhang are from SenseTime Research, China.
Chenhao Lin is from School of Cyber Science and Engineering,  Xi'an Jiaotong University, Xi'an, China.}
\thanks{Hongbin Sun and Zhanghui Kuang are co-first authors.}
\thanks{Corresponding author: Zhanghui Kuang (email: kuangzhh@gmail.com).}}

%
%

\markboth{Journal of \LaTeX\ Class Files,~Vol.~14, No.~8, August~2015}%
{Shell \MakeLowercase{\textit{et al.}}: Bare Demo of IEEEtran.cls for IEEE Journals}
%
\maketitle
\begin{abstract}
  	Key information extraction from document images is of paramount importance in office automation.
	Conventional template matching based approaches fail to generalize well to document images of unseen templates, and are not robust against text recognition errors.
	In this paper, we propose an end-to-end \textbf{S}patial \textbf{D}ual-\textbf{M}odality \textbf{G}raph \textbf{R}easoning method (SDMG-R) to extract key information from unstructured document images.
	We model document images as dual-modality graphs, nodes of which encode both the visual and textual features of detected text regions, and edges of which represent the spatial relations between neighboring text regions. The key information extraction is solved by iteratively propagating messages along graph edges and reasoning the categories of graph nodes.
	In order to roundly evaluate our proposed method as well as boost the future research, we release a new dataset named \textbf{WildReceipt}, which is collected and annotated tailored for the evaluation of key information extraction from document images of unseen templates in the wild.
	It contains 25 key information categories, a total of about 69000 text boxes, and is about 2 times larger than the existing public datasets. Extensive experiments validate that all information including visual features, textual features and spatial relations can benefit key information extraction.
	It has been shown that SDMG-R can effectively extract key information from document images of unseen templates, and obtain new state-of-the-art results on the recent popular benchmark SROIE and our WildReceipt.
	Our code and dataset will be publicly released.
\end{abstract}


%
\IEEEpeerreviewmaketitle

\section{Introduction}
%
%
%
%
\IEEEPARstart{E}{xtracting} key information from unstructured document images, such as historical documents, receipts, orders and credit notes, plays an important role in office automation including efficient archiving, compliance checking and so on. Conventional approaches~\cite{Schuster2013,rusinol2013field,cesarini2003analysis,medvet2011probabilistic} maintain a set of templates, each of which consists of key words and their layouts.
Although they can usually accurately extract key information from documents, they are not robust against the partial text recognition errors, which usually occurs.
To make matters worse, they cannot generalize to documents from unseen templates, which prohibits them from being used in many real application scenarios.

\begin{figure}
\center
    \begin{minipage}{0.8\linewidth}
    \centerline{ \includegraphics[width=\linewidth]{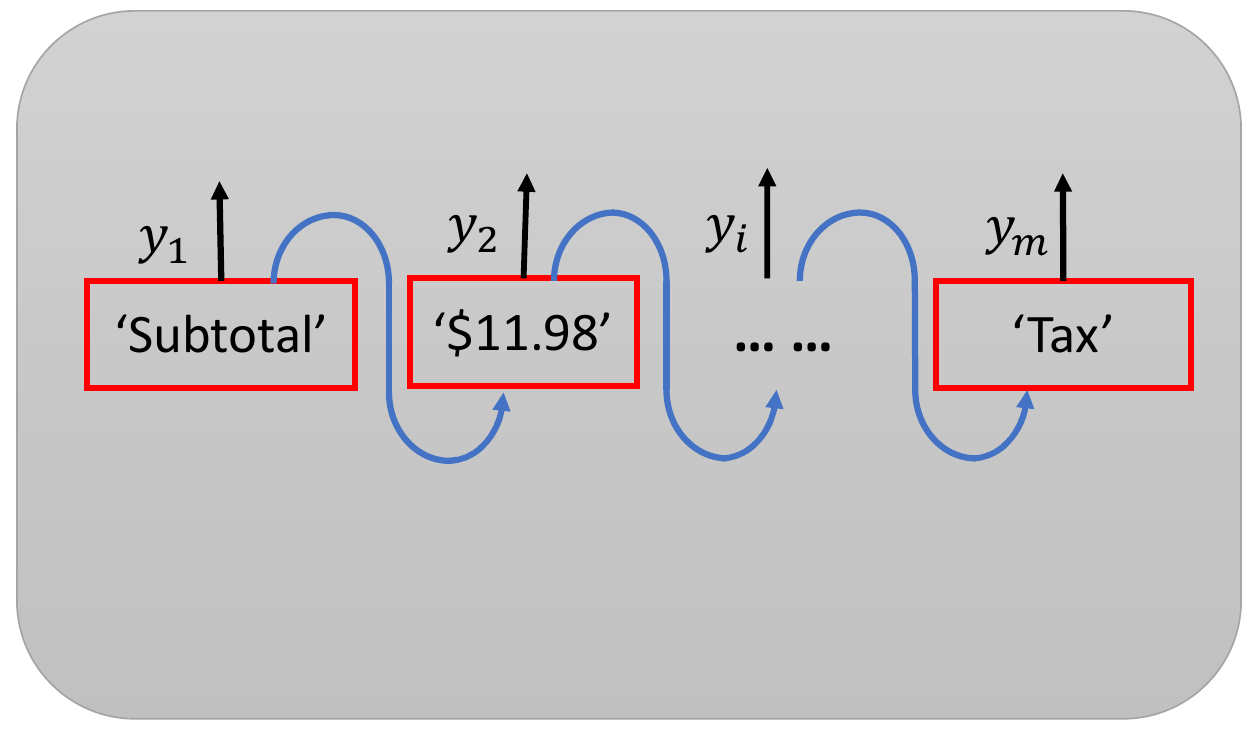}}
    \vfill
    \centerline{(a) NER }
    \end{minipage}
    \hfill
    \begin{minipage}{0.8\linewidth}
    \centerline{ \includegraphics[width=\linewidth]{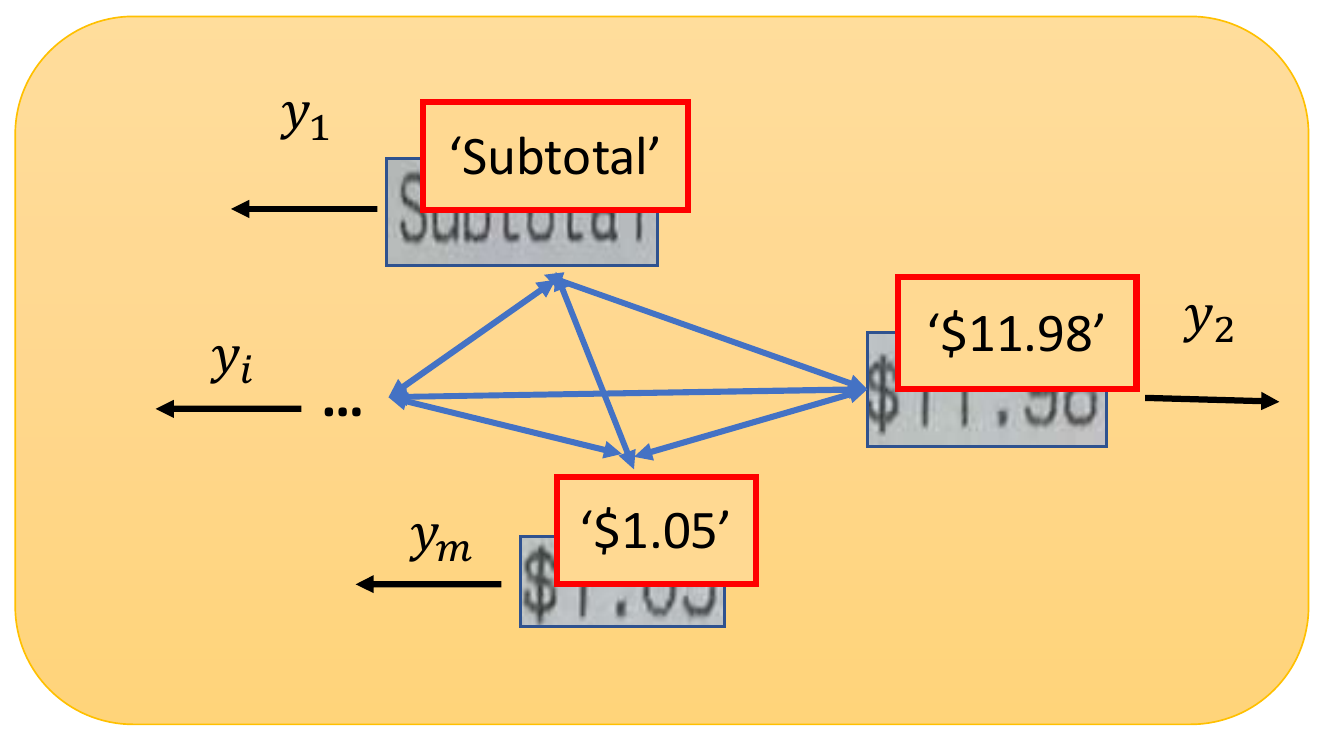}}
    \vfill
   \centerline{(b) SDMG-R}
   \end{minipage}
    \caption{Illustration of Named Entity Recognition (NER) and our proposed Spatial Dual-Modality Graph Reasoning (SDMG-R). NER models the relations between two text regions at the same horizontal line while our SDMG-R  between all text regions in the spatial neighborhood. Moreover, NER  use textual features only while SDMG-R both visual features extracted from image regions and textual ones extracted from texts.}
    \label{fig:introduction}
    \vspace{-6mm}
\end{figure}

In this paper, we target at key information extraction with a more challenging setting, where training set and test set have different templates.
CloudScan~\cite{palm2017cloudscan} modeled the key information extraction as Named Entity Recognition problem via concatenating texts as strings, which are classified as predefined categories such as order ID, invoice number and so on (see Figure~\ref{fig:introduction} (a)).
Although it can generalize to samples of unseen templates, it degrades greatly when lines are not aligned properly due to non-front image captures.
Moreover, it makes full use of pre-context and after-context only in the concatenated strings but not neighboring text regions which are not in the same line.
We believe that a robust key information extraction approach should be robust against image views, and utilize all context in the spatial neighborhood but not the same horizontal line only.

To this end, we propose an end-to-end Spatial Dual-Modality Graph Reasoning (SDMG-R) approach for key information extraction.
We model the unstructured document images as spatial dual-modality graphs with graph nodes as detected text boxes and graph edges as the spatial relations between these nodes (see Figure~\ref{fig:introduction} (b)).
Each node is associated with the textual and visual features which are learned via a recurrent neural network (RNN) and convolutional neural network (CNN) automatically.
Features of graph nodes are propagated iteratively along graph edges before classifying into pre-defined key information categories.
In this way, SDMG-R makes full use of spatial relations between detected text regions, and their dual modality features.
It is independent of document templates, and thus naturally has the potential to extract key information from document images of unseen templates.

Most of previous key information extraction approaches are evaluated on private data only due to the lack of public datasets.
Recently, a few datasets such as IEHHR~\cite{fornes2017icdar2017}, SROIE~\cite{sroie}, which target at key information extraction, have been emerging.
However, their training and test set share many templates, and thus they are unsuited to evaluate the generalization ability of key information extraction methods.
To this end, we build a new key information extraction benchmark dubbed WildReceipt.
It consists of 25 key information categories, totally about 50000 text boxes, which is about 2 times larger than SROIE.
The key information categories in WildReceipt are fine-grained. \textit{e.g.}, they contain ``Subtotal value'', ``Total value'' and ``Tax value'' categories, all of which are money amount, and it is difficult to distinguish with each other without context.
Different from previous scanned images as in SROIE, receipt images in WildReceipt are captured in the wild, most of them are from non-front views and with folds.
Therefore, it is more challenging and realistic than previous ones.

We extensively evaluate our proposed SDMG-R on SROIE and WildReceipt. It has been shown that the proposed approach outperforms previous methods with impressive margins. We investigate the factors of the effectiveness of the proposed approach, and find that both the spatial relations and the dual modality features benefit the key information extraction. 

The contributions of this paper are as follows:
\begin{quote}
	\begin{itemize}
		\item We propose an effective spatial dual-modality graph reasoning network (dubbed SDMG-R) for key information extraction. To the best of our knowledge, our SDMG-R is the first key information extraction approach which jointly reasons key information categories on textual and visual features of text boxes and their 2-dimensional spatial relationships.
		\item We annotate a new benchmark, named  WildReceipt, to facilitate the future research of key information extraction, which is fine-grained, and 2 times bigger than its competitors. It targets at evaluating key information extraction from document images of unseen templates captured in the wild, which is not explored in previous datasets.
		\item We validate the effectiveness of the proposed SDMG-R on two benchmarks, \textit{i.e.}, SROIE and WildReceipt. Our proposed approach outperforms state-of-the-art approaches with impressive margins.
	\end{itemize}
\end{quote}
\section{Related Work}
\textbf{Dataset.} Key information extraction has been attracting a large number of researchers from computer vision and multimedia filed.
However, most of them conducted experiments on private datasets.
Intellix~\cite{Schuster2013} was trained on 8000 and tested on 4000 scanned documents with 10 annotated semantic categories. CloudScan~\cite{palm2017cloudscan} conducted experiments on $326,471$ scanned UBL invoices. Later, CUTIE annotated $4,484$  Spanish receipt documents captured in the wild including taxi receipts, Meals Entertainment (ME) receipts, and hotel receipts, with 9 different key information classes. However, all the aforementioned datasets are unavailable publicly.
Recently, SROIE~\cite{sroie} consisting of 600 scanned receipts from ICDAR 2019 Robust Reading Challenge is released with 4 categories. Namely, store name, store address, date and total price. Its training set and its corresponding ground truth are publicly available. However, the test set ground truth is not released. We annotate its test set for evaluating key information extraction approaches on it in our experiments.
Our WildReceipt targets at facilitating the key information extraction research and will be publicly released.
It is about 2 times and 5 times bigger than SROIE in terms of the total image number and the key information category number respectively.
Different from previous datasets which contain value categories only, WildReceipt contains both key and value categories such as ``Total key'' category (\textit{e.g.}, ``Total:'' and ``Total'' ) and ``Total value'' category (\textit{e.g.}, ``\$10.5'' and ``\$20''). We empirically find that key category classification during training can boost the classification performance of value categories.
Our WildReceipt contains fine-grained categories such as ``Total value'', ``Subtotal value'', and ``Tax value'', all of which represent money amount.
 Moreover, it is captured in the wild, which is more challenging and wider applicable. The detailed comparison between the key information extraction datasets is conducted in Table~\ref{tab:com_dataset}.

\begin{table}[t]
	\scriptsize
	\begin{center}
		
		\caption{Comparison between different datasets. ``FG'' indicates fine-grained.} \label{tab:com_dataset}
		\begin{tabular}{c|c|c|c|c|c|c|c}
			\hline\hline
			Dataset&\#img&\#class&Source&Wild&FG&Key&Public \\ \hline
			Intellix&12,000&10&unknown&&&& \\
			CloudScan&326,471&8&invoice&&&&\\
			CUTIE&4,484&9&receipt&\Checkmark&&&\\
			SROIE&600&4&receipt&&&&\Checkmark\kern-1.5ex\raisebox{1ex}{\rotatebox[origin=c]{125}{\textbf{--}}}\\
			Ours&1768&25&receipt&\Checkmark&\Checkmark&\Checkmark&\Checkmark\\ \hline\hline
		\end{tabular}
	\end{center}
\end{table}

\begin{figure*}[t] 
	\centering
	\includegraphics[width=1.0\linewidth]{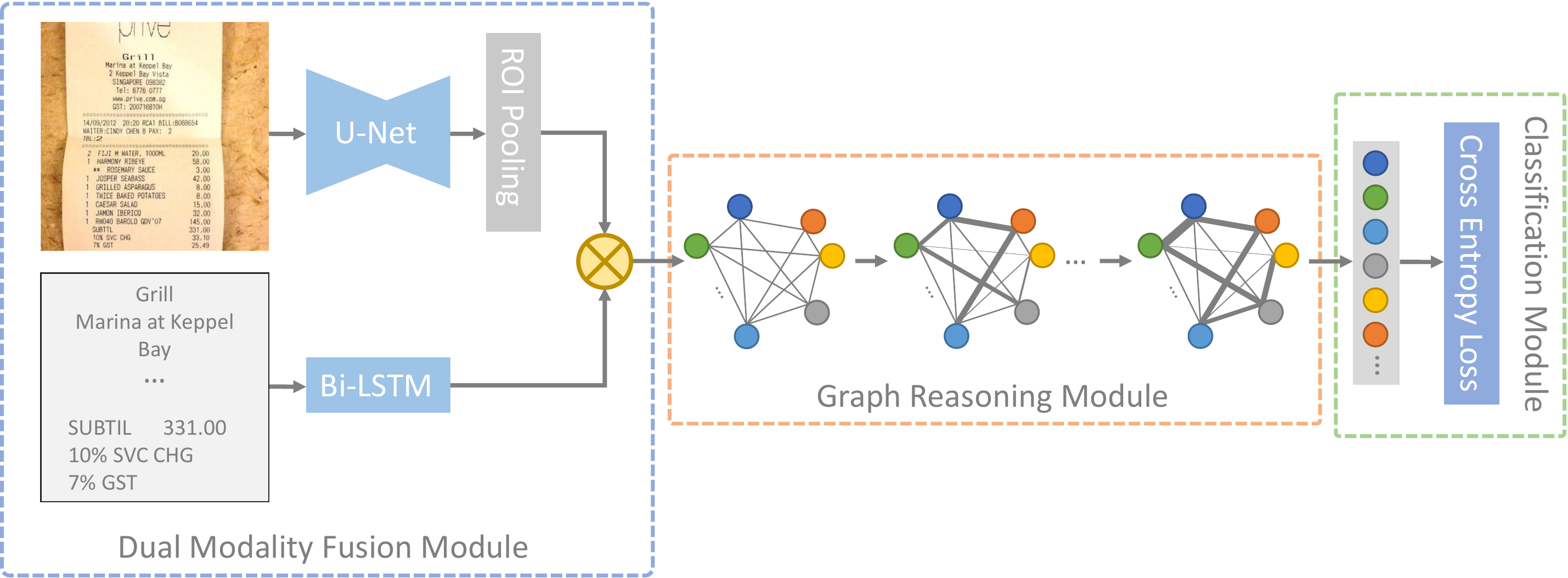}
	\caption{The overall architecture of the proposed SDMG-R model for key information extraction. Given one image, visual features $\{\mathbf{v_i}\}$ are extracted via U-Net and ROI-Pooling while textual features $\{\mathbf{t_i}\}$ are extracted via one Bi-LSTM. The modality features are fused by Kronecker product approximated by the block-diagonal tensor decomposition in the Dual Modality Fusion Module before being fed into the Graph Reasoning Module, where the node features are propagated and aggregated, and the edge weights are dynamically learned. The final node features are classified into one of key information categories in the classification module.}
	\label{fig:fig_overall}
\end{figure*}

\textbf{Key information extraction.}
Conventional approaches~\cite{Schuster2013,rusinol2013field,cesarini2003analysis,d2018field}
utilized template matching or rule based strategies, and thus performed poorly on documents of unseen templates.
Later, CloudScan~\cite{palm2017cloudscan} modeled the key information extraction problem as NER~\cite{peng2017cross,lample2016neural,chiu2016named,ma2016end}, and concatenated the entire invoice texts as one dimensional sequences without utilizing the two-dimensional spatial layout information.
Chargrid~\cite{Faddoul2018} encoded each document page as a two-dimensional grid of characters to conduct semantic segmentation.
It utilized two-dimensional spatial layout information with small neighborhood only,
and could not make fully use of nonlocal spatial relations between text regions with long distances due to limited effective receptive filed of convolution neural networks~\cite{effectivereceptivefield2016}. Recently, VRD~\cite{Liu2019} learned graph embedding to summarize the two-dimensional context of text segments in the document, which were further combined with text embedding for entity extraction from visually rich documents. Our proposed SDMG-R models documents as fully-connected graphs with text regions as nodes and two-dimensional spatial relations as edges. Different from VRD, SDMG-R learns both visual features and textual features of text regions, which leads to robustness against text recognition errors. Detailed comparison between recent approaches with our SDMG-R is given in Table~\ref{tab:tab_cmp_method}.

\begin{table}[t]
	\scriptsize
	\begin{center}
		\caption{Comparison between different key information extraction approaches in terms of whether utilizing 1D context, 2d context, nonlocal context and visual features of text regions.} \label{tab:tab_cmp_method}
		\begin{tabular}{c|c|c|c|c}
			\hline\hline
			Methods&1d context&2d context&Nonlocal context&Visual features  \\ \hline
CloudScan&\Checkmark&&& \\
Chargrid&\Checkmark&\Checkmark&& \\
VRD&\Checkmark&\Checkmark&\Checkmark& \\
SDMG-R&\Checkmark&\Checkmark&\Checkmark&\Checkmark
\\ \hline\hline
		\end{tabular}
	\end{center}
\end{table}



\textbf{Graph neural networks.}
Recently, integrating graphs with deep neural networks is an emerging topic in deep learning research.
A considerable amount of models have arisen for reasoning on graph-structured data at various tasks, such as classification of graphs \cite{duvenaud2015convolutional,dai2016discriminative,ni2016progressively}, 
classification of nodes of graphs \cite{kipf2018neural,kipf2016semi}, and modeling multi-agent interacting physical systems \cite{kipf2018neural,hoshen2017vain,sukhbaatar2016learning}. 
Graph neural networks have been widely used in many application such as human action recognition \cite{yan2018spatial,wang2018videos},  social relationship understanding \cite{Wang2018a},  object parsing \cite{Liang2016}, multi-label image recognition~\cite{Chen2018c}, visual question answer \cite{Teney2017}, and fashion retrieval~\cite{Kuang2019}.
These work create graphs via modeling the relations between image objects or regions.
In contrast,  we explore the use of graph to express spatial relations between detected text boxes which are encoded by visual-textual features, and apply it to the field of key information extraction. For each detected text box, it can automatically mine the useful layout structure in its neighborhood.

\textbf{Multi-modality fusion.}
Our work is also related to multi-modality fusion methods~\cite{Fukui2016,Kim2017,Ben-younes2017,Ben-younes2019}.
Most of them targets at visual question answering, visual grounding and visual relationship detection.
We are the first to investigate visual and textual modality fusion for key information extraction.

\section{Approach}
Given one document image $\mathbf{I}$ of size $H\times W$, together with detected text regions $\{\mathbf{r_i}\}$, where $\mathbf{r_i}=<x_i,y_i,h_i,w_i,s_i>$ with $(x_i,y_i)$, $h_i$, $w_i$, and $s_i$ being the top-left corner coordinate, the height, the width, and the recognized text string of $\mathbf{r_i}$ respectively, the key information extraction aims at classifying each detected text region $\mathbf{r_i}$ into one of a predefined category set $\mathcal{Y}$. We model the key information extraction as the graph node classification problem via jointly making full use of dual modality features. Namely, visual features and textual ones. Our proposed spatial dual-modality graph reasoning model consists of the dual-modality fusion module, the graph reasoning module and the classification module. Figure~\ref{fig:fig_overall} shows its overall architecture.

\subsection{Dual-Modality Fusion Module}
\label{sec:dual_modality_fusion_module}
Given one image with the text regions $\{\mathbf{r_i}\}$, we learn a feature vector $\mathbf{n}_i \in \mathcal{R}^{D_n}$ to represent each text  region $\mathbf{r_i}$ via the dual-modality fusion module.
The dual-modality fusion module is designed to effectively learn and compose the visual features and textual ones. We extract the visual feature $\mathbf{v_i}\in \mathcal{R}^{D_v}$ for $\mathbf{r_i}$ by RoI Pooling~\cite{fastrcnn2015} with its rectangle $(x_i,y_i,h_i,w_i)$ on the output  feature maps of the last layer of one CNN feature extractor.
In our experiments, we use U-Net~\cite{RFB15a} to instantiate the CNN feature extractor.
Besides, we extract the textual feature for $\mathbf{r_i}$ by designing a char-level Bi-LSTM~\cite{ma2016end}. Specifically,  we first represent each char in $s_i$ as a one-hot vector $\mathbf{e_i^j}\in \mathcal{R}^{D_c}$   with dimension $D_c$ being the cardinality of the char dictionary. $\mathbf{e_i^j}$ is then projected into a lower dimensional space and finally sequentially fed into the Bi-LSTM module to obtain the textual representation $\mathbf{t_i}\in \mathbf{D}^t$ for the text region $r_i$. Formally, we have
\begin{equation}
\mathbf{t_i}=\text{Bi-LSTM} (\mathbf{W_s}\mathbf{e_i^j}),
\end{equation}
where $\mathbf{W_s}\in \mathcal{R}^{D_s\times D_c}$ is the projection matrix of textual one-hot vectors.
We fuse the visual features and textual features via modeling the interactions between all possible visual and textual feature dimensional pairs, which are easily obtained with Kronecker product as follows:
\begin{equation}
\mathbf{n}_i=\mathbf{P}({\mathbf{t_i}}\otimes{\mathbf{v_i}}),
\label{eq:eq_kp}
\end{equation}
$\otimes$ is the Kronecker product operation.  $\mathbf{P}\in \mathcal{R}^{D_n\times (D_v\times D_t)}$ is one learnable linear transformation and $\mathbf{n_i} \in \mathcal{R}^{D_n}$ is the fused feature.  For simplicity, we ignore the bias term in our paper.
The number of learnable parameters in Equation (\ref{eq:eq_kp}) grows linearly with the dimension of the visual features, that of the textual ones, and that of the fused representations, which results in heavy memory and computation overheads. To reduce the memory and computation complexity, we first reformulate Equation (\ref{eq:eq_kp}) into tensor form:
\begin{equation}
\mathbf{n}_i=\mathbf{T}\times_1{\mathbf{t_i}^{\intercal}}\times_2{\mathbf{v_i}^{\intercal}},
\label{eq:eq_tensor}
\end{equation}
where $\mathbf{T}\in \mathcal{R}^{D_t\times D_v\times D_n}$ is one tensor via reshaping $\mathbf{P}$ in Equation (\ref{eq:eq_kp}), and $\times_j$ are the mode-$j$ product.
$\mathbf{x}^{\intercal}$ indicates the transpose of $\mathbf{x}$. We then introduce the block tensor decomposition ~\cite{YeWLCZCX18,Ben-younes2019} to decompose $\mathbf{T}$ as follows:
\begin{equation}
\mathbf{T}=\mathbf{C_b}\times_1 \mathbf{P_t} \times_2 \mathbf{P_v} \times_3 \mathbf{P_n},
\label{eq:eq_decomposation}
\end{equation}
where $\mathbf{C_b}\in \mathcal{R}^{D_t^bR\times D_v^bR\times D_n^bR}$ is the block-diagonal core tensor with $R$ being the block number and
$D_t^b\times D_v^b\times D_n^b$ being the block size, $\mathbf{P_t} \in \mathcal{R}^{D_t\times D_t^b R}$, $\mathbf{P_v} \in \mathcal{R}^{D_v\times D_v^bR}$ and $\mathbf{P_n} \in \mathcal{R}^{D_n^bR\times D_n}$. Usually,  we set $D_t\gg D_t^b$, $D_v\gg D_v^b$, $D_n\gg D_n^b$ and $R$ to one small constant. Thus, the parameter number decomposed tensor in Equation (\ref{eq:eq_decomposation}) is greatly smaller than that of original tensor in Equation (\ref{eq:eq_tensor}). \textit{i.e.}, $R(D_t^b\times D_v^b \times D_n^b)+D_tD_t^bR+D_vD_v^bR+D_nD_n^bR\ll D_vD_tD_n$ as shown in our experiments.



We also implement alternative fusion schemes in our experiments for comparison.

\textbf{LinearSum.} The visual features $\mathbf{x_v}$ and textual features $\mathbf{x_t}$ are linearly projected into one common space $\mathcal{R}^{D_n}$ via one three-layer MLP, and then element-wise added as the fused representation of $\mathbf{n_i}$.

\textbf{ConcatMLP.} The visual features $\mathbf{x_v}$ and textual features $\mathbf{x_t}$ are concatenated, followed by one three-layer MLP.

\subsection{Graph Reasoning Module}
We model the document images as graphs $\mathcal{G}=(\mathcal{N},\mathcal{E})$,  where
$\mathcal{N}=\{\mathbf{n}_i\}$ with  $\mathbf{n_i}$ being the feature vector of the text node $\mathbf{r_i}$, and $\mathcal{E}=\{e_{ij}\}$ with $e_{ij}$ being the edge weight between the node $\mathbf{r_i}$ and the node $\mathbf{r_j}$.

We encode the spatial relation $e_{ij}\in \mathcal{R}$ between $\mathbf{r_i}$ and $\mathbf{r_j}$ via one dynamic attention mechanism.  We first define the spatial relation between node $\mathbf{r_i}$ and $\mathbf{r_j}$ as follows:
\begin{align}
\Delta x_{ij}&=x_j-x_i,\\
\Delta y_{ij}&=y_j-y_i, \\
\mathbf{r_{ij}^p}&=\frac{\Delta x_{ij}}{d}\mathbin\Vert\frac{\Delta y_{ij}}{d}, \label{eq:eq_distance_normalization}\\
\mathbf{r_{ij}^s}&=\frac{w_{i}}{h_i}\mathbin\Vert\frac{h_j}{h_i}\mathbin\Vert \frac{w_j}{h_i},\label{eq:eq_relative_shape}\\
\mathbf{r_{ij}}&=\mathbf{r}_{ij}^p\mathbin\Vert\mathbf{r}_{ij}^s,
\end{align}
where $\Delta x_{ij}$, and $\Delta y_{ij}$ are the horizontal distance and the vertical one between the two text boxes $\mathbf{r_i}$ and $\mathbf{r_j}$ respectively. $d$ is one normalization constant, and $\mathbin\Vert$ is the concatenation operation.  The spatial  position relation between two text boxes plays a critical role in key information extraction. $\mathbf{r_{ij}^p}$ encodes the relative spatial  position distance between node $\mathbf{r_i}$ and $\mathbf{r_j}$.
The first term and latter two ones in Equation~(\ref{eq:eq_relative_shape}) encode the aspect ratio of $\mathbf{r_i}$ and relative shape information respectively.

\begin{figure*}[t] 
	\centering
	\includegraphics[width=0.9\linewidth]{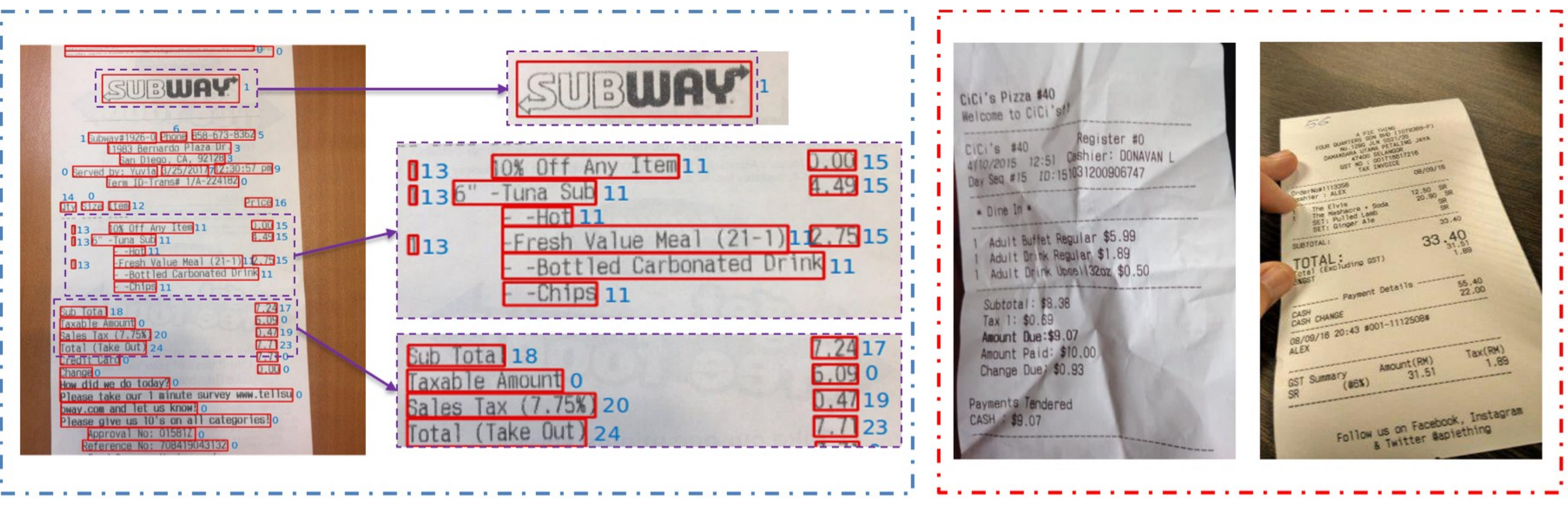}
	\caption{Annotations and samples of WildReceipt. The left shows the annotated text bounding boxes (red) with their corresponding key information categories (blue); The right shows one receipt sample with folds, and one non-front sample in WildReceipt (best viewed in color).}
	\label{fig:wrong_image}
\end{figure*}

Inspired by~\cite{gan2018graph}, we  embed the spatial information between text boxes into the edge weight $e_{ij}\ \in \mathcal{E}$ as follows:

\begin{align}
\mathbf{e_{ij}^{'}}&=N_{l_2}(\mathbf{E}\mathbf{r}_{ij}), \label{eq:eq_eij_p}\\
\mathbf{e_{ij}}&=\mathbf{n_i}\mathbin\Vert\mathbf{e_{ij}^{'}}\mathbin\Vert\mathbf{n_j},\label{eq:eq_tri}\\
e_{ij} &=  M(\mathbf{e_{ij}}), \label{eq:eq_edge_mlp}
\end{align}
where $\mathbf{E}\in \mathcal{R}^{D_e\times 5}$ is one linear transformation which embeds the spatial relation information $\mathbf{r_{ij}}$ into a $D_e$-dimensional representation. $N_{l_2}$ is the $l_2$ normalization operation, which is introduced to stabilize the training procedure.
$\mathbf{e_{ij}}\in \mathcal{R}^{(D_e+2*D_n)}$ is the concatenated representations of $\mathbf{n_i}$, $\mathbf{n_j}$ and the normalized spatial relation embedding. $M$ is one MLP which transforms $\mathbf{e_{ij}}$ into the scalar $e_{ij}$.

\textbf{Graph reasoning.} We iteratively refine the features $\{\mathbf{n_i}\}$  of the proposed spatial dual-modality graph $L$ times as follows:
\begin{equation}
\mathbf{n_i^{l+1}} = \mathbf{n_i^{l}} +\sigma(\mathbf{W^l}(\sum_{j\ne i}\alpha_{ij}^l \mathbf{e_{ij}^l})),
\end{equation}
where $\mathbf{n_i^{l}} \in \mathbf{R}^{ D_n}$ indicates the feature of the $i^{th}$ graph node  at time step $l$.
$\alpha_{ij}^l$ is the normalized graph edge weight at time step $l$.
$\mathbf{W^l}\in \mathbf{R}^{D_n\times (D_e+2D_n)}$ is a linear transformation at time step $l$.
$\mathbf{e_{ij}^l}$ is the concatenated representation of  $\mathbf{n_i^l}$, $\mathbf{n_j^l}$ and the normalized spatial relation embedding
at time step $l$ as described in Equation~(\ref{eq:eq_tri}).
$\sigma$ is the ReLU nonlinear activation.
$\alpha_{ij}^l$ is the learnable normalized edge weight between node $i$ and $j$ at time step $l$.
It is given by
\begin{align}
\alpha_{ij}^l&=\frac{ \text{exp}(e_{ij})}{\sum_{k\ne i}\text{exp}(e_{ik})}, \label{eq:eq_affinity}
\end{align}
From Equation~(\ref{eq:eq_affinity}), the edge weights of the proposed graph $\mathcal{G}$ change dynamically during reasoning from one iteration to another.

\subsection{Loss}
The final output $\mathbf{n}^L$ of the iterative reasoning module is fed to the classification module to classify each text region into one of key information categories.
Formally, our loss is defined as

\begin{equation}
\begin{aligned}
Loss= \sum_i\text{CE}(\mathbf{n}_i^L, y_i)
\end{aligned}
\end{equation}
where $y_i\in \mathcal{Y}$ is the key information category ground truth.

\begin{table*}[t] 
	\scriptsize
	\begin{center}
		\caption{Statistics text bounding boxes of our WildReceipt and SROIE. To represent the category name more concisely, we use the abbreviation of words in names to represent its category, \textit{e.g.}, `Str nm' and `Prod item' denote the store name and the product item, respectively.}
		\label{tab:cat-table}
			\resizebox{\textwidth}{12mm}{
				\begin{tabular}{c|c|c|c|c|c|c|c|c|c|c|c|c|c|c|c|c|c|c|c|c|c|c|c|c|c}
					\hline
					\hline
					\rotatebox{90}{Category}&\rotatebox{90}{Str nm key}&\rotatebox{90}{Str nm value}&\rotatebox{90}{Str addr key}&\rotatebox{90}{Str addr value}&\rotatebox{90}{Tel key}&\rotatebox{90}{Tel value}&\rotatebox{90}{Date key}&\rotatebox{90}{Date value}& \rotatebox{90}{Time key}&\rotatebox{90}{Time value}&\rotatebox{90}{Prod item key}&\rotatebox{90}{Prod item value}&
					\rotatebox{90}{Prod qty key}&\rotatebox{90}{Prod qty value}&
					\rotatebox{90}{Prod price key}&\rotatebox{90}{Prod price value}&\rotatebox{90}{Subtotal key}&\rotatebox{90}{Subtotal value}&
					\rotatebox{90}{Tax key}&\rotatebox{90}{Tax value}&\rotatebox{90}{Tips key}&\rotatebox{90}{Tips value}&
					\rotatebox{90}{Total key}&\rotatebox{90}{Total value}&\rotatebox{90}{Others} \\\hline
					
					Ours&3&1682&2&2347&452&1067&342&1673&193&1525&368&8634&334&5272&373&8401&1269&1352&1413&1415&151&165&1783&2136&26623\\\hline
					
					SROIE&0&663&0&1530&0&0&0&638&0&0&0&0&0&0&0&0&0&0&0&0&0&0&0&589&26108\\\hline
					\hline						
		\end{tabular}}
	\end{center}
\end{table*}

\section{WildReceipt}
\subsection{Data Collection}
We selected receipts to benchmark key information extraction as SROIE~\cite{sroie} due to the following reasons:
(1) receipts are anonymous, and suitable for public release without private information leak; (2) receipts are of varied templates since different companies usually have different templates. Thus, it is suitable for evaluating key information extraction from document images of unseen templates;
(3) receipts are widely available and easy to collect; (4) extracting key information from receipts have many applications such as bookkeeping, and reimbursement.

We collected and annotated WildReceipt in the following procedure.
\begin{itemize}
	\item Data collection. We searched receipt images on search engines with related key-words, such as receipt, invoice and so on. We downloaded about 4300 document images.
	\item Data cleaning. We removed images which have multiple receipts inside, are not receipts, unreadable, incomplete, or non-English manually.
	\item Data annotation. We first labelled the text bounding boxes and their corresponding texts, and then labelled each bounding box to one of 25 key information categories (see Figure~\ref{fig:wrong_image}). These annotations were done by 6 experts.
\end{itemize}
The receipt images in WildReceipt we selected are captured in the wild. They are of non-front views and possibly with folds as shown in Figure~\ref{fig:wrong_image}. Therefore, WildReceipt is much more challenging than previous key information extraction benchmarks which
focus on scanned documents only.

\subsection{Statistics}
The WildReceipt dataset consists of 1740 receipt images, 68975 text bounding boxes. Each image has average about 39 text bounding boxes.
Table~\ref{tab:cat-table} lists the annotation numbers of all 25 key information categories.
In the 25 key information categories, 12 categories are keys and 12 categories are their corresponding values, and 1 category indicates others.
As there are many variants for one type of key, \textit{e.g.}, ``Address'', ``address'', and ``Add.'' all indicate the key category ``Str addr key''.
We believe that accurately identifying key categories can benefit greatly key information extraction, which is validated in our experiments. WildReceipt is 2 times and 3 times larger than SROIE~\cite{sroie} in terms of the image number and the category number. Besides, it contains fine-grained key information categories. \textit{e.g.}, ``Product price value'', ``Tax value'', ``Tips value'' and ``Total value'' all are related with the amount of money, and difficult to distinguish with each other by their own textual or visual features without context information.


\subsection{Evaluation Protocol}
We randomly sampled 1268 and  472 images without replacement for training and test respectively.
During sampling, we made sure these two sets  had different templates according to store names and near-duplicated image retrieval~\cite{SivicZ03}.
In this way, the templates in test set are unseen in the training set. Therefore, WildReceipt is suitable for evaluating
key information extraction from document images of unseen templates.
Table~\ref{tab:tab_evaluation} lists the statistics for the training set and the test set in WildReceipt.

Performances on WildReceipt are evaluated by $F_1$ score as~\cite{sroie}.
The averaged $F_1$ score over $12$ value categories is finally reported.
WildReceipt will be publicly released to facilitate future research and fair comparison on key information extraction.

\begin{table}[t] 
	\small
	\begin{center}
		\caption{Statistics of training and test sets in WildReceipt.}
		\label{tab:tab_evaluation}
			\begin{tabular}{c|c|c}
				\hline
				\hline
				Dataset                     & Training     &  Test   \\ \hline
				Images                      & 1268         & 472 \\ \hline
				Templates                   & 847          & 345 \\\hline
				Total boxes                 & 49377        & 19598  \\\hline
				\hline
		\end{tabular}
	\end{center}
\end{table}
\section{Experiments}
In this section, the proposed approach SDMG-R is extensively evaluated on SROIE and WildReceipt.
We first introduce the implementation details. Then, SDMG-R is compared with state-of-the-art key information
extraction approaches quantitatively.
Finally, we investigate the effectiveness of each component of our proposed method by ablation study.

\subsection{Implementation Details}
Our implementation is based on PyTorch. Our models are trained on 1 NVIDIA Titan X GPUs with 12 GB memory.
During training, we randomly crop images with probability 0.5 while keeping all text boxes not cutting out. During test, we do not crop images. In both training and test, all images are resized to $512 \times 512$, and their text boxes are resized proportionally before being fed into the network.
The whole network is trained from scratch with default initializer of PyTorch using Adam optimizer~\cite{kingma2014adam}.
We use a batch size of 4 during training. Maximum epoch number is set to $60$.
The learning rate is set to $10^{-3}$ initially. It is decreased via $10\times$ after 40 and 50 epochs.

The cardinality of our dictionary is 91 (\textit{i.e.}, $D_c=91$).
It contains 0-9 digital, a-z and A-Z letters, and special characters which are greatly related to
key information categories. They are ``/'', ``$\backslash$'', ``.'', ``\$'', ``\euro{}'', ``\textlira'', ``\textyen'',
``:'', ``-'', ``*'', ``\#'', ``('', ``)'', ``\%'', ``@'', ``!'', ``''', ``\&'', ``='', ``>'', ``+'', ``"'', ``$\times$'', ``?'', ``<'', ``['', ``]'', and ``\_''.  All other characters in texts are set to one token ``unkown''.  The one-hot char encoding vectors are projected to a 32-dimensional space (\textit{i.e.}, $D_s=32$).
The dimension of the hidden vector of Bi-LSTM $D_t$ is set to $256$.
As for visual modality, we adopt the U-Net~\cite{RFB15a} as our visual feature extractor, and extract visual features on its last convolutional output feature maps, followed by a dimensional reduction to $256$. Thus, we have $D_v=256$.
In the block tensor decomposition module, we set $D_t^b = D_v^b = D_n^b =52$ and $R=20$. We set the graph node feature representation dimension to $256$ (\textit{i.e.}, $D_n=256$).  The normalization constant is set to $10$ (\textit{i.e.},$d=10$) in Equation (\ref{eq:eq_distance_normalization}). The 5-dimensional edge features are embedded into one 256-dimensional space (\textit{i.e.}, $D_e=256$). The MLP ($M$ in Equation (\ref{eq:eq_edge_mlp})) is of one two layers with one ReLU between them. Its hidden dimension is $256$. The graph reasoning iteration number is set to 2 (\textit{i.e.}, $L=2$.) except otherwise noted.

\subsection{Comparison with State-of-the-art Methods}
We compare our proposed SDMG-R with  two state-of-the-art approaches and their variants. Specially, we evaluate the following methods:
\begin{itemize}
	\item Chargrid~\cite{Faddoul2018}.  It models documents as two-dimensional grids of characters, which are fed into a fully convolutional neural network to predict segmentation masks.
	\item Chargrid-UNet.  For fair comparison, we also use U-Net as Chargrid's backbone while keeping other unchanged. We name Chargrid with this setting as Chargrid-UNet.
	\item VRD~\cite{Liu2019}.  It models documents with text bounding boxes as graphs, which are then fed into one CRF. 	
\end{itemize}

\begin{table*}[tp]
	\centering
	\fontsize{8}{8}\selectfont
	\caption{Comparison with state-of-the-art methods on WildReceipt in terms of $F_1$ score ($\%$).}
	\label{tab:comp_sota_wild_receipt_gt}
	\begin{tabular}{c|c|c|c|c|c|c|c|c|c|c|c|c|c}
		\hline
		\hline
		Method & Str nm & Str addr&Tel &Date &Time &Prod item & Prod qty & Prod price & Subtotal & Tax & Tips & Total &Avg\cr
		\hline
		Chargrid     &78.4&79.0&86.2&90.0&87.0&92.0&93.6&92.0&68.0&68.1&20.5&70.1&76.9 \cr
		Chargrid-UNet&80.6&82.0&85.2&90.6&86.3&89.6&94.2&92.0&66.9&68.6&39.0 &72.6&79.0 \cr
		VRD           &78.0&82.8&92.2&95.1&94.9&91.8&95.8&95.6&82.4&84.2&56.0 &79.5&85.7 \cr \hline
		SDMG-R     &\textbf{79.8}&\textbf{85.7}&\textbf{94.0}&\textbf{95.7}&94.7&\textbf{93.9}&95.6&\textbf{97.1}&\textbf{87.9}&\textbf{89.5}&\textbf{67.9}&\textbf{82.4}&\textbf{88.7} \cr\hline
		\hline
	\end{tabular}
\end{table*}
\begin{table*}[tp]
	\centering
	\fontsize{8}{8}\selectfont
	\caption{Comparison with state-of-the-art methods on WildReceipt with ground truth text boxes and recognized texts in terms of $F_1$ score ($\%$).}
	\label{tab:performance_comparison_ocr}
	\setlength{\tabcolsep}{0.9mm}
	\begin{tabular}{c|c|c|c|c|c|c|c|c|c|c|c|c|c}
		\hline
		\hline
		Method & Str nm & Str addr&Tel &Date &Time &Prod item & Prod qty & Prod price & Subtotal & Tax & Tips & Total &Avg\cr
		\hline
		Chargrid  &69.1&72.3&80.7&85.6&81.7&84.2&88.6&90.9&65.9&63.6&5.8&68.2&71.4 \cr
		Chargrid-UNet    &69.5&74.5&79.1&85.6&83.4&85.3&86.9&89.2&62.8&64.5&12.5 &65.7&71.6 \cr
		VRD           &66.1&72.1&86.9&92.7&92.7&88.4&88.3&93.9&77.8&80.2&44.8 &74.2&79.8 \cr \hline
		SDMG-R     &\textbf{75.4}&\textbf{78.7}&\textbf{90.0}&92.0&92.2&\textbf{88.7}&\textbf{93.0}&\textbf{94.1}&\textbf{82.1}&\textbf{82.0}&\textbf{45.2}&\textbf{75.7}&\textbf{82.4} \cr\hline
		\hline
	\end{tabular}
\end{table*}

\begin{table*}[tp]
	\centering
	\fontsize{8}{8}\selectfont
	\caption{Comparison with state-of-the-art methods on WildReceipt with detected text boxes and recognized texts in terms of $F_1$ score ($\%$).}
	\label{tab:performance_comparison_ocr_detect}
	\begin{tabular}{c|c|c|c|c|c|c|c|c|c|c|c|c|c}
		\hline
		\hline
		Method & Str nm & Str addr&Tel &Date &Time &Prod item & Prod qty & Prod price & Subtotal & Tax & Tips & Total &Avg.\cr
		\hline
		Chargrid  &65.8&67.9&65.2&56.2&59.6&76.8&42.9&87.9&61.8&54.7&18.7&62.1&60.0 \cr
		Chargrid-UNet    &69.7&69.4&69.5&56.4&60.2&77.3&36.9&86.0&56.9&53.4&6.4 &61.8&58.7 \cr
		VRD           &67.9&77.9&80.7&89.1&90.5&82.3&65.4&91.7&77.6&82.3&44.8 &75.7&77.2 \cr \hline
		SDMG-R     &\textbf{72.6}&\textbf{79.8}&80.3&89.0&88.7&\textbf{86.8}&\textbf{80.3}&\textbf{92.3}&\textbf{78.0}&79.8&\textbf{50.0}&73.4&\textbf{79.3} \cr\hline
		\hline
	\end{tabular}
\end{table*}

\begin{table}[tp]	
	\centering
	\small
	\caption{Comparison with state of the art methods on SROIE in terms of $F_1$ score ($\%$).}
	\label{tab:performance_comparison}
	\begin{tabular}{c|c}
		\hline
		\hline
		Method & SROIE  \cr
		\hline
		Chargrid        &80.9\cr           
		Chargrid-UNet        &80.8 \cr
		VRD      &84.9 \cr\hline               
		SDMG-R          &{\bf 87.1}\cr\hline
		\hline
	\end{tabular}
\end{table}

We compare our proposed method with its counterparts in Table ~\ref{tab:comp_sota_wild_receipt_gt}. It has been
shown that our SDMG-R outperforms all its competitors with impressive margins. Specifically, SDMG-R achieves 11.8\%, 9.7\%,  and 3.0\% absolute improvements
in terms of $F_1$ score averaged on 12 value categories on WildReceipt compared with Chargrid, Chargrid-UNet, and VRD respectively.
Moreover, SDMG-R achieves best $F_1$ score for 10  out of 12 categories. Our SDMG-R is greatly superior than Chargrid-UNet. We believe it is because of the long range dependence between texts learned via graphs. Compared with VRD, the performance gain of the SDMG-R attributes to our proposed U-Net based visual modality and Kronecker product based modality fusion. For the categories ``Time'' and ``Prod qty'', our proposed SDMG-R and VRD are comparable.

Since in real applications, text boxes and texts are usually obtained by OCR engines, which might introduce text detection and recognition errors. To evaluate how those errors affect the performance of the key information extraction, we employ Google OCR API\footnote{https://cloud.google.com/vision/docs/ocr} to detect and recognize texts of WildReceipt. For each detected text box, we label its key information category as that of the ground truth text region of maximum IOU with it. We compare our SDMG-R with state-of-the-art methods when texts are recognized using the OCR engine given ground truth text boxes in Table \ref{tab:performance_comparison_ocr}.
Again, our proposed SDMG-R achieves the best averaged $F_1$ score. Moreover, it obviously outperforms its competitors in 10 out of 12 categories. Comparing Table~\ref{tab:comp_sota_wild_receipt_gt} and Table~\ref{tab:performance_comparison_ocr}, we  observe that there are about $6.3\%$ perform drop ($88.7\%$ v.s. $82.4\%$) in terms of averaged $F_1$ score if texts are recognized by the OCR engine. It is reasonable as
some of texts, especially, characters which are closely related to some specific key information categories such as ``\$'', ``\euro{}'', ``\textlira'', ``\textyen'' are misrecognized via the OCR engine, which results in noisy signals and poor discriminative representations.
To move forward, we compare our method with other methods in the case that both text boxes and texts are predicted by the OCR engine in Table~\ref{tab:performance_comparison_ocr_detect}. It has been shown that our proposed SDMG-R outperforms Chargird, Chargird-UNet and VRD with impressive margins. Note that there exists mismatching between detected text boxes and ground truth boxes. \textit{e.g.,} one detected text boxes might overlap with multiple ground truth ones or one ground truth text box might overlap with multiple detected ones. Directly matching text boxes with ground truth ones with maximum IOU might introduce noisy signals, which results in further perform drop. However, our method is still superior than its counterparts, which validates its robustness against noises.

We also compare our method with other start-of-the-art approaches on the dataset SROIE in Table~\ref{tab:performance_comparison}. Similar to WildReceipt, our SDMG-R obviously performs  better than others. Specially, it absolutely improves the $F_1$ scores of Chargrid, Chargrid-UNet, and VRD by $6.2\%$, $6.3\%$ and $2.2\%$ respectively. It has demonstrated the superiority of our SDMG-R on scanned document images.



\begin{table}[tp]	
	\centering
	\small
	\caption{Effectiveness of components of SDMG-R on WildReceipt in terms of the averaged $F_1$ score ($\%$).}
	\label{tab:result2}
	\begin{tabular}{c|c}
		\hline
		\hline
		Method & $F_1$  \cr
		\hline
		w/o textual features        &80.1 \cr           
		w/o visual  features        & 86.4\cr   \hline        
		w/o spatial relation      &81.8 \cr               
		w/o graph reasoning       &77.2 \cr \hline          
		w/o key category classification&84.3 \cr \hline
		SDMG-R          &{\bf 88.7}\cr\hline
		\hline
	\end{tabular}
\end{table}

\begin{table}[t] 
	\small
	\begin{center}
		\caption{Performance comparison of SDMG with various modality fusion methods on WildReceipt in terms of the averaged $F_1$ score ($\%$).}
		\label{tab:graph_layers_evaluation}
		\begin{tabular}{c|c|c}
		\hline
		\hline
		Method & Settings&$F_1$  \cr
		\hline
		 ConcatMLP &$D_h=256$&    0.860 \cr           
		 ConcatMLP &$D_h=512$&    \underline{0.867} \cr           
		 ConcatMLP &$D_h=768$&    \underline{0.867} \cr           
		 ConcatMLP &$D_h=1024$&   0.854 \cr  \hline         
		
		 LinearSum &$D_h=512$&0.859 \cr           
		 LinearSum &$D_h=768$&\underline{0.861} \cr           
		 LinearSum &$D_h=1024$&0.845 \cr   \hline        
		 Our &$D_t^b=D_v^b=D_n^b = 26, R=20$&0.859\cr
		 Our &$D_t^b=D_v^b=D_n^b = 52, R=20$&\underline{\textbf{0.887}}\cr
		Our &$D_t^b=D_v^b=D_n^b = 78, R=20$&0.857\cr \hline
		\hline
	\end{tabular}
		
	\end{center}
\end{table}

\subsection{Ablation Studies}
We perform detailed ablation studies on WildReceipt to investigate the effectiveness of each component of our proposed SDMG-R.

\textbf{Effects of visual and textual features.}
In Table~\ref{tab:result2}, SDMG-R decreases absolutely by  $8.6\%$ on WildReceipt in terms of $F_1$ score when without textual features.
Similarly, it decreases absolutely by $2.3\%$  when without visual features. It has been shown that both textual and visual features, especially, textual features, contribute the key information extraction greatly.

\textbf{Effects of spatial relation.} To cancel out the spatial relation, we set the edge weights $\mathbf{e_{ij}^{'}}=\mathbf{0}$ in Equation (\ref{eq:eq_eij_p}) for all graph node pairs $(i,j)$. SDMG-R decreases its $F_1$ score to $81.8\%$ on WildReceipt. We have observed that spatial relations between two text boxes play an important role in key information extraction and can boost its performance obviously.

\textbf{Effects of graph reasoning.} Without graph reasoning, we directly conduct classification over the fused visual and textual features, resulting in great performance degradation. Namely, absolute $11.5\%$ $F_1$ score drop on WildReceipt. It suggests that message propagation between text regions can refine their representations so that they can be correctly classified into their corresponding key information categories.

\textbf{Effects of key category classification.} In our WildReceipt dataset, we also annotate key categories such as ``Str nm key'', although only the information of value categories needs to be extracted in real application scenarios. However, we experimentally find that key category classification can help the value category classification. As shown in Table~\ref{tab:result2}, without it, our SDMG-R decreases absolutely by 4.4\%.

\textbf{Effects of graph reasoning iteration number.} Our SDMG-R obtains the averaged $F_1$ score of $84.9\%$, $88.7\%$ and $87.6\%$ when the graph reasoning iteration number $L$ is set to $1$, $2$ and $3$ respectively. It achieves the best performance when $L=2$, and is overfitted when $L>2$. We set $L=2$ in our experiments.

\textbf{Effects of dual modality fusion module.}
Dual modality fusion module is the core component to fuse visual and textual features. We compare our module with its counterparts LinearSum and ConcatMLP as described in Section~\ref{sec:dual_modality_fusion_module}.  For fair comparison, we enumerate the hidden dimension ($D_h$) of the MLP in LinearSum and ConcatMLP, and the block size ({$D_t^b$, $D_v^b$, $D_n^b$}) and the block num ($R$) in our proposed dual modality fusion module, and report their corresponding results in Table~\ref{tab:graph_layers_evaluation}. We can observe that ConcatMLP and LinearSum achieve their best results with $D_h=512$ (or $D_h=768$), and $D_h=768$ respectively while our method with $D_t^b=D_v^b=D_n^b=52$ and $R=20$. It has been shown that our proposed dual modality fusion module is very effective, and obviously outperforms its alternative methods. Namely, LinerSum and ConcatMLP.

\begin{figure}
    \begin{minipage}{0.48\linewidth}
    \centerline{ \includegraphics[width=\linewidth]{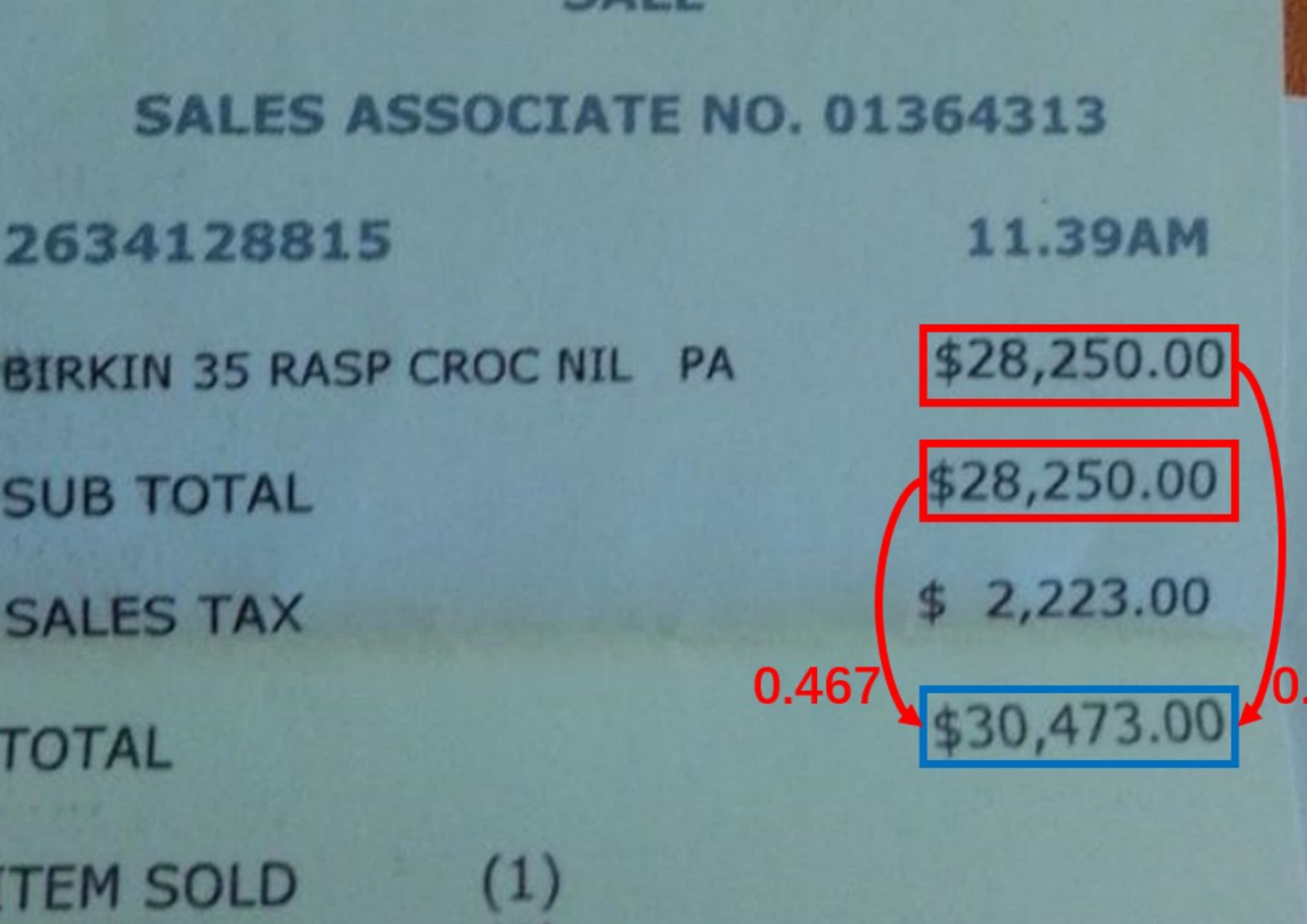}}
    \end{minipage}
    \begin{minipage}{0.48\linewidth}
    \centerline{ \includegraphics[width=\linewidth]{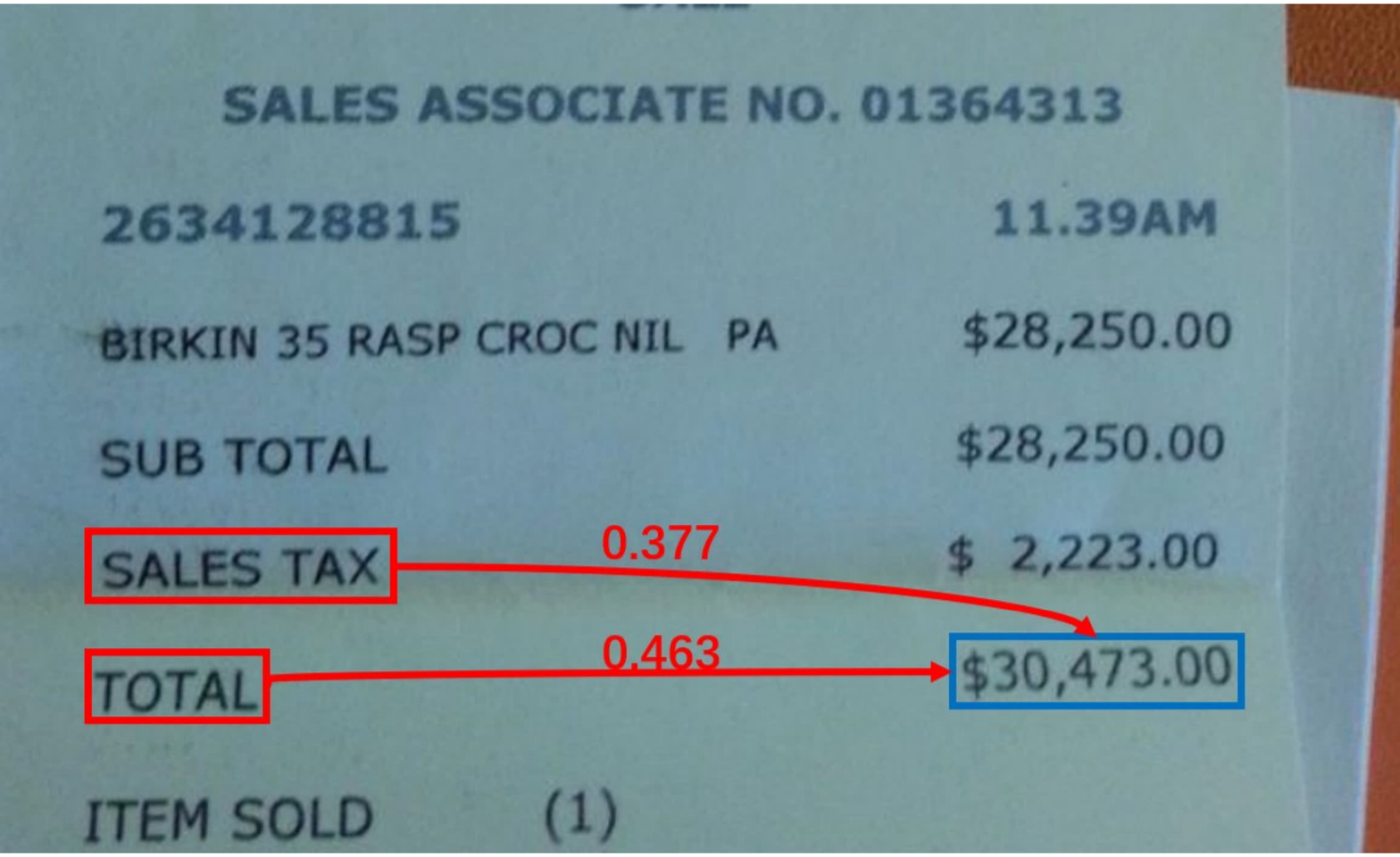}}
   \end{minipage}

\begin{minipage}{0.48\linewidth}
    \centerline{ \includegraphics[width=\linewidth]{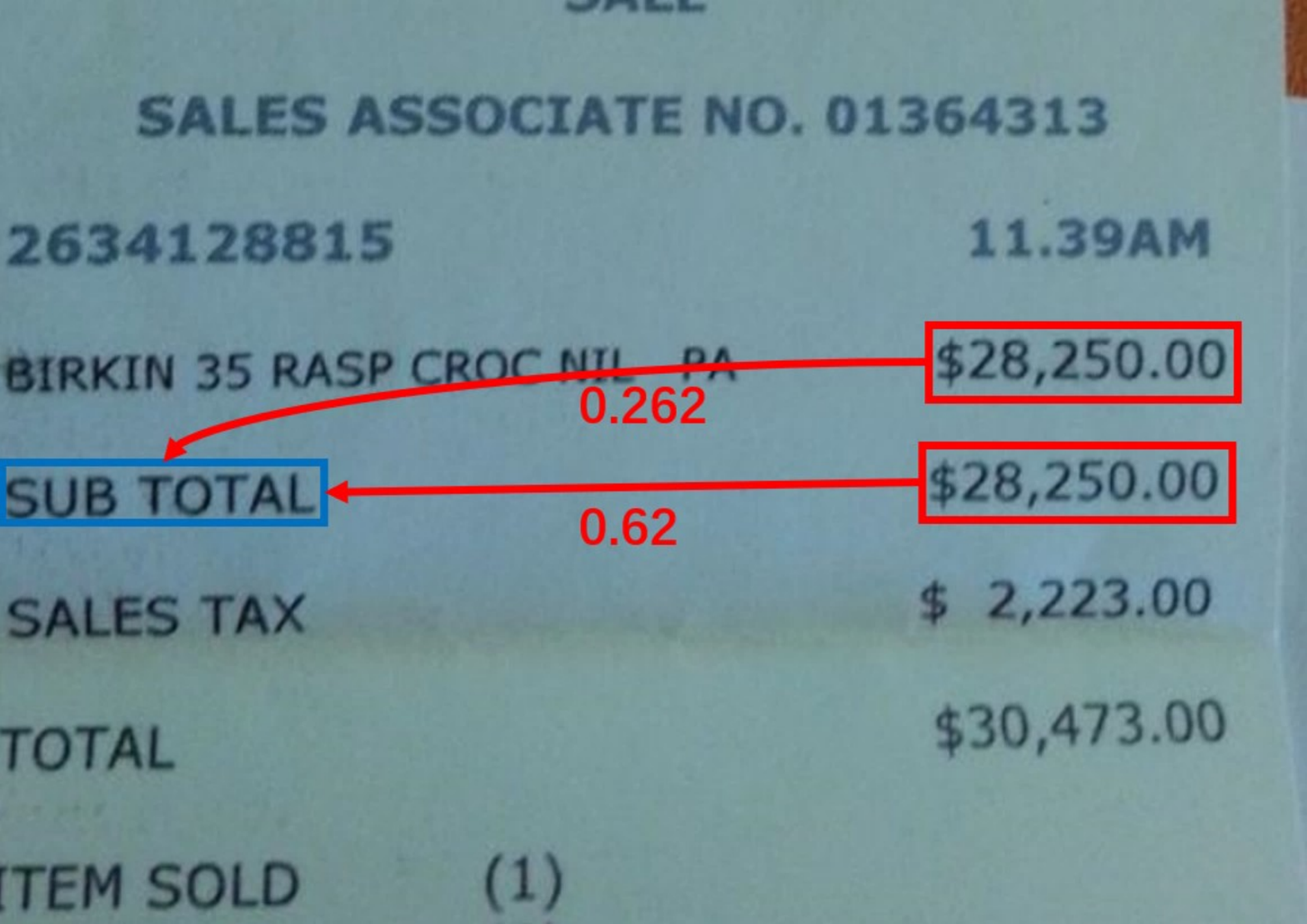}}
    \end{minipage}
    \begin{minipage}{0.48\linewidth}
    \centerline{ \includegraphics[width=\linewidth]{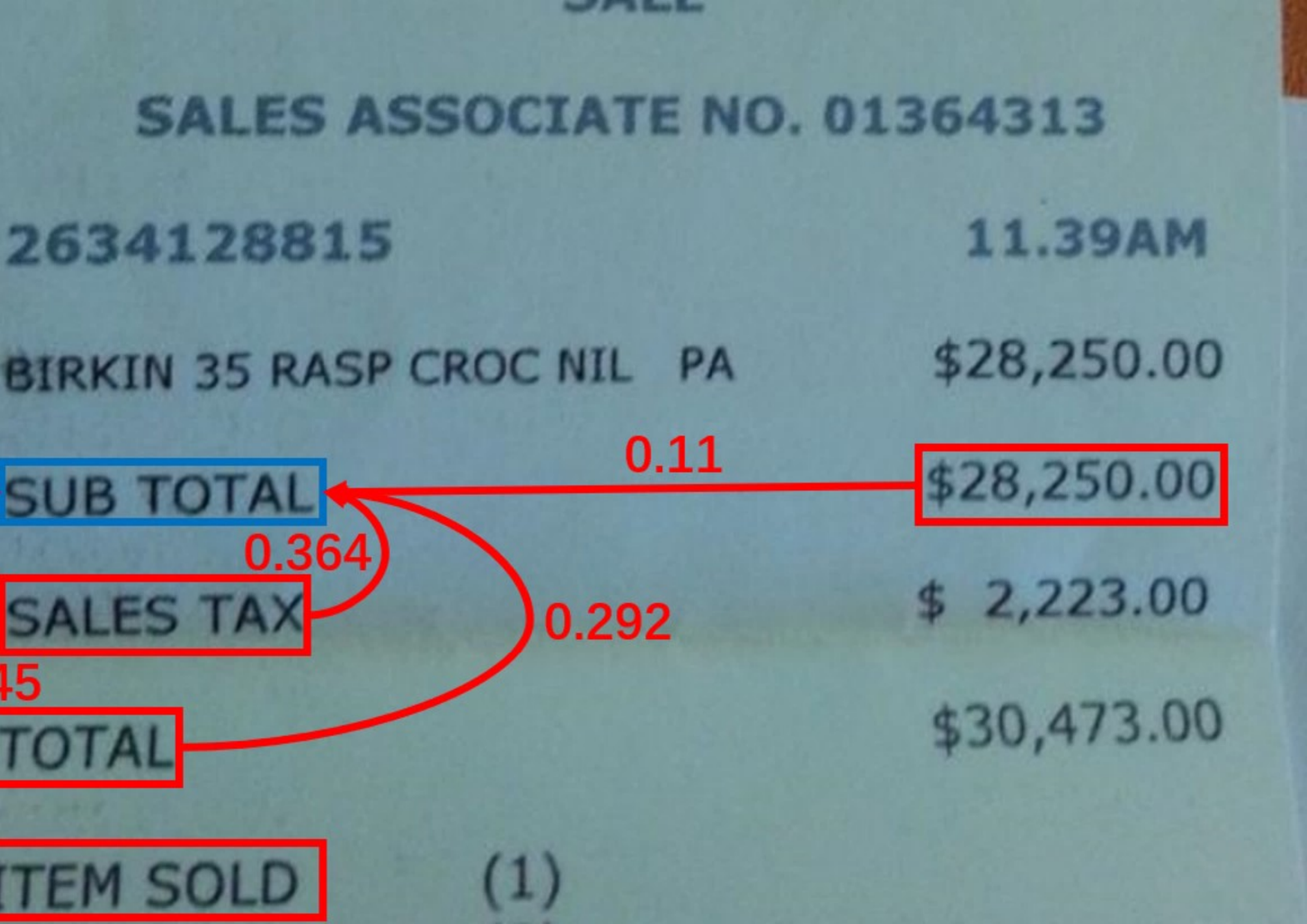}}
   \end{minipage}

   \begin{minipage}{0.48\linewidth}
    \centerline{ \includegraphics[width=\linewidth]{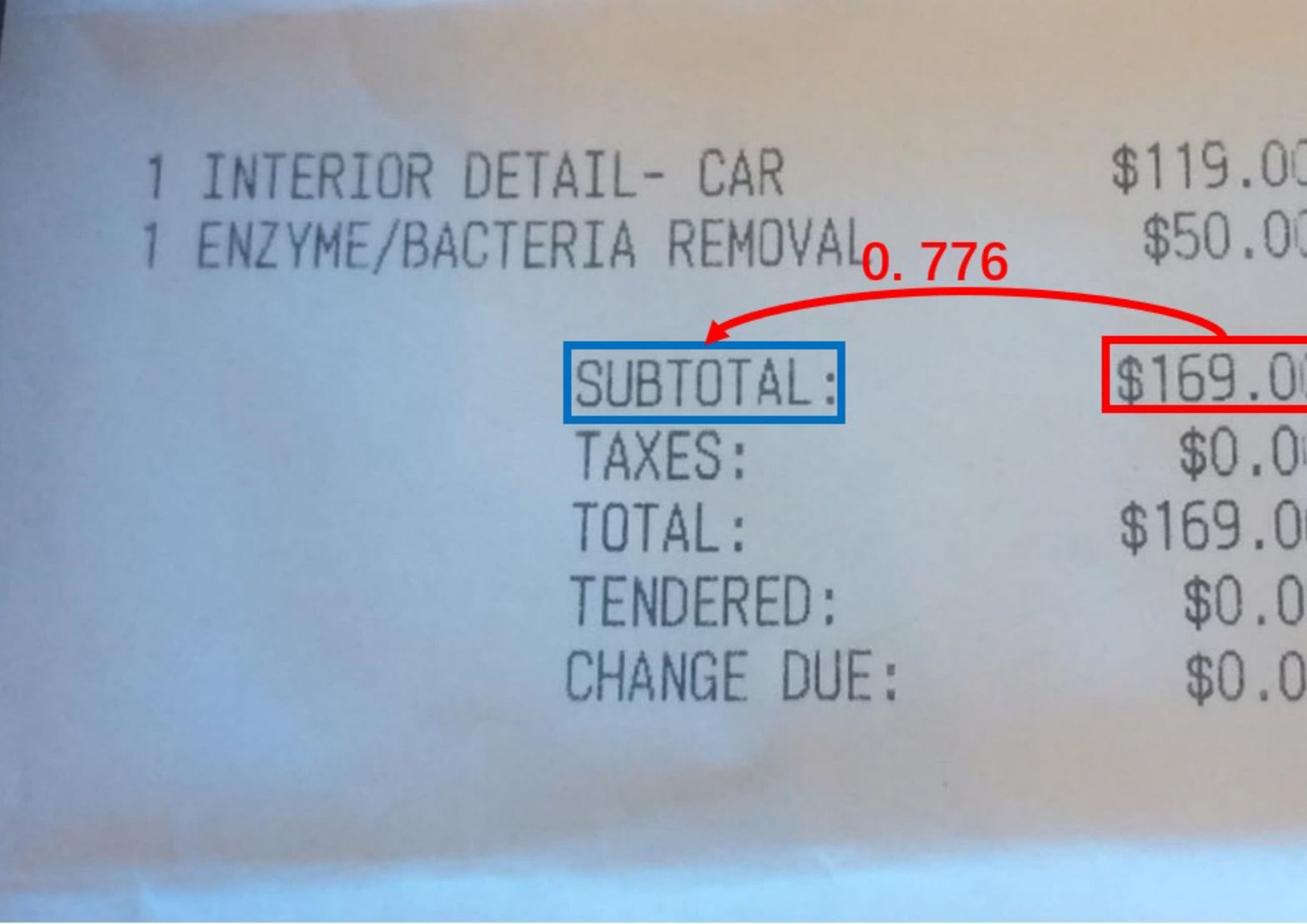}}
    \centerline{First GCL}
    \end{minipage}
    \begin{minipage}{0.48\linewidth}
    \centerline{ \includegraphics[width=\linewidth]{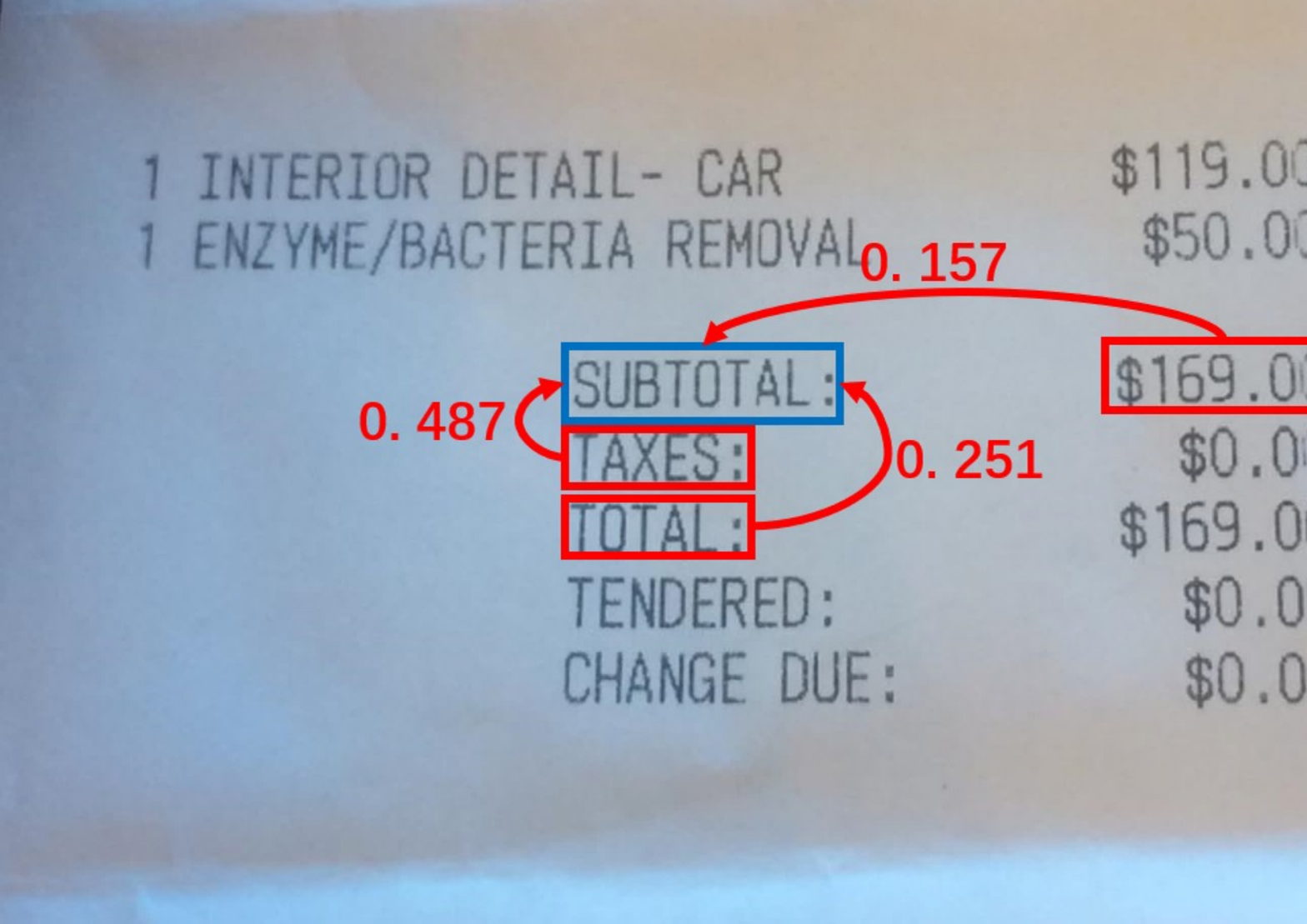}}
   \centerline{Second GCL}
   \end{minipage}
   \caption{Visualization of the learned dynamic weight $e_{ij}$ between text regions $i$ and $j$. Each row shows one text region (blue rectangle) and its related regions (red rectangles) of one receipt image. The first column shows the learned weights in the first Graph Convolution Layer (GCL) in our graph reasoning module while the second column shows those in the second GCL. We visualize the weights (the red numbers), which are bigger than 0.1, of edges (the red directed curves) incoming to one node only for clarity (best viewed in color).}
    \label{fig:graph_result}
    \vspace{-6mm}
\end{figure}

\subsection{Visualization}
To better understand how our SDMG-R learns the spatial relations between the text regions, we visualize the learned edge weights $e_{ij}$  in Figure~\ref{fig:graph_result}. Interestingly, it can highlight the edges between two semantically-related text regions even they are
with long spatial distances. \textit{e.g.},  the edges between ``Total value'', ``Subtotal value'' and ``Prod item value'' (top left), those between ``Total value'', ``Total key'', and ``Tax key'' (top right), those between ``Subtotal value'' and ``Subtotal key'' (bottom left), and those between ``Subtotal key'', ``Tax key'', ``Total key'' and ``Subtotal value'' (bottom right). Compared with the first GCL (the left column), the second GCL can learn more helpful spatial relations to identity the key information categories of text regions. \textit{e.g.}, ``TOTAL'' on the left of ``\$30473.00'' highly indicates  ``\$30473.00'' is one instance of ``Total value'' in the top right subfigure. ``TOTAL:'' and ``TAXES:'' under the ``SUBTOTAL:'' indicate ``SUBTOTAL:'' is one instance of ``Subtotal key'' in the bottom right subfigure.
\section{Conclusions}
In this paper, we have proposed a novel spatial dual-modality graph reasoning model (termed SDMG-R) for key information extraction from unstructured documents. We have introduced Kronecker product approximated via the block diagonal tensor decomposition to fuse the visual and textual features. SDMG-R naturally learns spatial relations between text regions via dynamical attentions in its graph reasoning module. We have validated the effectiveness of each component of the proposed SDMG-G by extensive experiments. Moreover, a new large key information extraction dataset, named WildReceipt, has been annotated to  evaluate the model performance of the key information extraction  on document of unseen templates. It is fine grained and captured in the wild, and thus more challenging and realistic than previous public datasets. It will be publicly released  for facilitating future research. Experimental results on both SROIE and our WildReceipt databases have shown that our proposed SDMG-R consistently outperforms start-of-the-art key information extraction methods with impressive margins.

\ifCLASSOPTIONcaptionsoff
  \newpage
\fi



%
\bibliographystyle{IEEEtran}
\bibliography{IEEEabrv,sample-base}

\begin{thebibliography}{10}
\providecommand{\url}[1]{#1}
\csname url@samestyle\endcsname
\providecommand{\newblock}{\relax}
\providecommand{\bibinfo}[2]{#2}
\providecommand{\BIBentrySTDinterwordspacing}{\spaceskip=0pt\relax}
\providecommand{\BIBentryALTinterwordstretchfactor}{4}
\providecommand{\BIBentryALTinterwordspacing}{\spaceskip=\fontdimen2\font plus
\BIBentryALTinterwordstretchfactor\fontdimen3\font minus
  \fontdimen4\font\relax}
\providecommand{\BIBforeignlanguage}[2]{{%
\expandafter\ifx\csname l@#1\endcsname\relax
\typeout{** WARNING: IEEEtran.bst: No hyphenation pattern has been}%
\typeout{** loaded for the language `#1'. Using the pattern for}%
\typeout{** the default language instead.}%
\else
\language=\csname l@#1\endcsname
\fi
#2}}
\providecommand{\BIBdecl}{\relax}
\BIBdecl

\bibitem{Schuster2013}
D.~Schuster, K.~Muthmann, D.~Esser, A.~Schill, M.~Berger, C.~Weidling,
  K.~Aliyev, and A.~Hofmeier, ``{Intellix - End-User Trained Information
  Extraction for Document Archiving},'' in \emph{ICDAR}, 2013.

\bibitem{rusinol2013field}
M.~{Rusiñol}, T.~{Benkhelfallah}, and V.~P. {dAndecy}, ``Field extraction from
  administrative documents by incremental structural templates,'' in \emph{2013
  12th International Conference on Document Analysis and Recognition}, 2013,
  pp. 1100--1104.

\bibitem{cesarini2003analysis}
F.~Cesarini, E.~Francesconi, M.~Gori, and G.~Soda, ``Analysis and understanding
  of multi-class invoices,'' \emph{Document Analysis and Recognition}, vol.~6,
  no.~2, pp. 102--114, 2003.

\bibitem{medvet2011probabilistic}
E.~Medvet, A.~Bartoli, and G.~Davanzo, ``A probabilistic approach to printed
  document understanding,'' \emph{International Journal on Document Analysis
  and Recognition (IJDAR)}, vol.~14, no.~4, pp. 335--347, 2011.

\bibitem{palm2017cloudscan}
R.~B. Palm, O.~Winther, and F.~Laws, ``Cloudscan-a configuration-free invoice
  analysis system using recurrent neural networks,'' in \emph{ICDAR}, vol.~1,
  2017, pp. 406--413.

\bibitem{fornes2017icdar2017}
A.~Forn{\'e}s, V.~Romero, A.~Bar{\'o}, J.~I. Toledo, J.~A. S{\'a}nchez,
  E.~Vidal, and J.~Llad{\'o}s, ``Icdar2017 competition on information
  extraction in historical handwritten records,'' in \emph{ICDAR}, vol.~1,
  2017, pp. 1389--1394.

\bibitem{sroie}
Z.~Huang, K.~Chen, J.~He, X.~Bai, D.~Karatzas, S.~Lu, and C.~V. Jawahar,
  ``{ICDAR 2019 Robust Reading Challenge on Scanned Receipts OCR and
  Information Extraction},'' 2019.

\bibitem{d2018field}
V.~P. {D'}Andecy, E.~Hartmann, and M.~Rusinol, ``Field extraction by hybrid
  incremental and a-priori structural templates,'' in \emph{2018 13th IAPR
  International Workshop on Document Analysis Systems (DAS)}, 2018, pp.
  251--256.

\bibitem{peng2017cross}
N.~Peng, H.~Poon, C.~Quirk, K.~Toutanova, and W.-t. Yih, ``Cross-sentence n-ary
  relation extraction with graph lstms,'' \emph{Transactions of the Association
  for Computational Linguistics}, vol.~5, pp. 101--115, 2017.

\bibitem{lample2016neural}
G.~Lample, M.~Ballesteros, S.~Subramanian, K.~Kawakami, and C.~Dyer, ``Neural
  architectures for named entity recognition,'' \emph{arXiv preprint
  arXiv:1603.01360}, 2016.

\bibitem{chiu2016named}
J.~P. Chiu and E.~Nichols, ``Named entity recognition with bidirectional
  lstm-cnns,'' \emph{Transactions of the Association for Computational
  Linguistics}, vol.~4, pp. 357--370, 2016.

\bibitem{ma2016end}
X.~Ma and E.~Hovy, ``End-to-end sequence labeling via bi-directional
  lstm-cnns-crf,'' \emph{arXiv preprint arXiv:1603.01354}, 2016.

\bibitem{Faddoul2018}
A.~R.~K. Faddoul, C.~R.~C. Guder, S.~Brarda, S.~Bickel, J.~H{\"{o}}hne, and
  J.~Baptiste, ``{Chargrid: Towards Understanding 2D Documents},'' in
  \emph{EMNLP}, 2018.

\bibitem{effectivereceptivefield2016}
W.~Luo, Y.~Li, R.~Urtasun, and R.~Zemel, ``Understanding the effective
  receptive field in deep convolutional neural networks,'' in \emph{Advances in
  Neural Information Processing Systems 29}, D.~D. Lee, M.~Sugiyama, U.~V.
  Luxburg, I.~Guyon, and R.~Garnett, Eds., 2016, pp. 4898--4906.

\bibitem{Liu2019}
X.~Liu, F.~Gao, Q.~Zhang, and H.~Zhao, ``{Graph Convolution for Multimodal
  Information Extraction from Visually Rich Documents},'' in \emph{NAACL},
  2019, pp. 32--39.

\bibitem{duvenaud2015convolutional}
D.~K. Duvenaud, D.~Maclaurin, J.~Iparraguirre, R.~Bombarell, T.~Hirzel,
  A.~Aspuru-Guzik, and R.~P. Adams, ``Convolutional networks on graphs for
  learning molecular fingerprints,'' in \emph{NeurIPS}, 2015, pp. 2224--2232.

\bibitem{dai2016discriminative}
H.~Dai, B.~Dai, and L.~Song, ``Discriminative embeddings of latent variable
  models for structured data,'' in \emph{ICML}, 2016, pp. 2702--2711.

\bibitem{ni2016progressively}
B.~Ni, X.~Yang, and S.~Gao, ``Progressively parsing interactional objects for
  fine grained action detection,'' in \emph{CVPR}, 2016, pp. 1020--1028.

\bibitem{kipf2018neural}
T.~Kipf, E.~Fetaya, K.-C. Wang, M.~Welling, and R.~Zemel, ``Neural relational
  inference for interacting systems,'' \emph{arXiv preprint arXiv:1802.04687},
  2018.

\bibitem{kipf2016semi}
T.~N. Kipf and M.~Welling, ``Semi-supervised classification with graph
  convolutional networks,'' \emph{arXiv preprint arXiv:1609.02907}, 2016.

\bibitem{hoshen2017vain}
Y.~Hoshen, ``Vain: Attentional multi-agent predictive modeling,'' in
  \emph{NeurIPS}, 2017, pp. 2701--2711.

\bibitem{sukhbaatar2016learning}
S.~Sukhbaatar, A.~Szlam, and R.~Fergus, ``Learning multiagent communication
  with backpropagation,'' in \emph{NeurIPS}, 2016, pp. 2244--2252.

\bibitem{yan2018spatial}
S.~Yan, Y.~Xiong, and D.~Lin, ``Spatial temporal graph convolutional networks
  for skeleton-based action recognition,'' in \emph{AAAI}, 2018.

\bibitem{wang2018videos}
X.~Wang and A.~Gupta, ``Videos as space-time region graphs,'' in \emph{ECCV},
  2018, pp. 399--417.

\bibitem{Wang2018a}
\BIBentryALTinterwordspacing
Z.~Wang, T.~Chen, J.~Ren, W.~Yu, H.~Cheng, and L.~Lin, ``{Deep Reasoning with
  Knowledge Graph for Social Relationship Understanding},'' in \emph{IJCAI},
  2018. [Online]. Available: \url{http://arxiv.org/abs/1807.00504}
\BIBentrySTDinterwordspacing

\bibitem{Liang2016}
X.~Liang, X.~Shen, J.~Feng, L.~Lin, and S.~Yan, ``{Semantic Object Parsing with
  Graph LSTM},'' in \emph{ECCV}, 2016.

\bibitem{Chen2018c}
T.~Chen, Z.~Wang, G.~Li, and L.~Lin, ``{Recurrent Attentional Reinforcement
  Learning for Multi-label Image Recognition},'' in \emph{AAAI}, 2018.

\bibitem{Teney2017}
D.~Teney, L.~Liu, and A.~v.~D. Hengel, ``{Graph-Structured Representations for
  Visual Question Answering},'' in \emph{CVPR}, 2017, pp. 1--9.

\bibitem{Kuang2019}
Z.~Kuang, Y.~Gao, G.~Li, P.~Luo, Y.~Chen, L.~Lin, and W.~Zhang, ``{Fashion
  Retrieval via Graph Reasoning Networks on a Similarity Pyramid},'' in
  \emph{ICCV}, 2019.

\bibitem{Fukui2016}
A.~Fukui, D.~H. Park, D.~Yang, A.~Rohrbach, T.~Darrell, and M.~Rohrbach,
  ``{Multimodal Compact Bilinear Pooling for Visual Question Answering and
  Visual Grounding},'' in \emph{EMNLP}, 2016, pp. 457--468.

\bibitem{Kim2017}
J.~H. Kim, K.~W. On, W.~Lim, J.~Kim, J.~W. Ha, and B.~T. Zhang, ``{Hadamard
  Product for Low-rank Bilinear Pooling},'' in \emph{ICLR}, 2017, pp. 1--14.

\bibitem{Ben-younes2017}
H.~Ben-younes, M.~Cord, and N.~Thome, ``{MUTAN: Multimodal Tucker Fusion for
  VQA},'' in \emph{ICCV}, 2017, pp. 2612--2620.

\bibitem{Ben-younes2019}
H.~Ben-younes, R.~Cadene, N.~Thome, and M.~Cord, ``{BLOCK: Bilinear
  Superdiagonal Fusion for Visual Question Answering and Visual Relationship
  Detection},'' in \emph{AAAI}, vol.~33, 2019, pp. 8102--8109.

\bibitem{fastrcnn2015}
R.~Girshick, ``Fast r-cnn,'' in \emph{ICCV}, 2015.

\bibitem{RFB15a}
O.~Ronneberger, P.Fischer, and T.~Brox, ``U-net: Convolutional networks for
  biomedical image segmentation,'' in \emph{Medical Image Computing and
  Computer-Assisted Intervention (MICCAI)}, vol. 9351, 2015, pp. 234--241.

\bibitem{YeWLCZCX18}
J.~Ye, L.~Wang, G.~Li, D.~Chen, S.~Zhe, X.~Chu, and Z.~Xu, ``Learning compact
  recurrent neural networks with block-term tensor decomposition,'' in
  \emph{{CVPR}}, 2018, pp. 9378--9387.

\bibitem{gan2018graph}
P.~VeliAkoviA, G.~Cucurull, A.~Casanova, A.~Romero, P.~LiA, and Y.~Bengio,
  ``Graph attention networks,'' in \emph{International Conference on Learning
  Representations}, 2018.

\bibitem{SivicZ03}
J.~Sivic and A.~Zisserman, ``Video google: A text retrieval approach to object
  matching in videos.'' in \emph{ICCV}, 2003, pp. 1470--1477.

\bibitem{kingma2014adam}
D.~P. Kingma and J.~Ba, ``Adam: A method for stochastic optimization,''
  \emph{arXiv preprint arXiv:1412.6980}, 2014.

\end{thebibliography}
\end{document}